\documentclass{article} % For LaTeX2e
\usepackage{tencent_ailab_tech_report}
\usepackage[colorlinks = true,
            linkcolor = blue,
            urlcolor  = blue,
            citecolor = blue,
            anchorcolor = blue]{hyperref}   
\usepackage{microtype}
\usepackage{hyperref}
\usepackage{url}
\usepackage{booktabs}
\usepackage{enumitem}
\usepackage{multicol}
\usepackage{multirow}
\usepackage{CJKutf8}
\usepackage{amsmath}
\usepackage{siunitx}
\usepackage{floatflt}
\usepackage{wrapfig}
\usepackage{graphicx}
\usepackage{booktabs}
\usepackage{wrapfig}
\usepackage{authblk}
\usepackage{lipsum}

\usepackage{algorithm}
\usepackage{algorithmicx}
\usepackage{algpseudocode}
\usepackage{microtype}
\usepackage{graphicx}
\usepackage{subfigure}
\usepackage{multirow}
\usepackage{booktabs} % for professional tables
\usepackage{pifont}  % 为了 \XSolidBrush
\usepackage{graphicx}  % 用于插入图像
\usepackage{subcaption} % 用于创建子图
\usepackage{hyperref}

\usepackage{amssymb} % 引入amssymb包

\algnewcommand{\LeftComment}[1]{\Statex \(\triangleright\) #1}

\usepackage{array}
\usepackage{amsmath}
\usepackage{amssymb}
\usepackage{mathtools}
\usepackage{amsthm}
\usepackage{arydshln}
\usepackage[capitalize,noabbrev]{cleveref}
\usepackage{adjustbox} % 引入adjustbox包
\usepackage{enumitem}

\theoremstyle{plain}

\theoremstyle{definition}

\theoremstyle{remark}

\usepackage[textsize=tiny]{todonotes}

\definecolor{nred}{RGB}{196, 38, 11}
\definecolor{ngreen}{RGB}{18, 141, 21}
\definecolor{nblue}{RGB}{41, 52, 190}
\definecolor{hzw}{RGB}{223, 97, 76}
\definecolor{lt}{RGB}{54, 89, 170}

\newcommand{\ignore}[1]{}

\colmfinalcopy

\title{Do NOT Think That Much for 2+3=? On the Overthinking of o1-Like LLMs}

\author[ ]{Xingyu Chen\thanks{Equal Contribution. The work was done when Xingyu and Zhiwei were interning at Tencent AI Lab.}~~$^{,1,2}$}
\author[ ]{Jiahao Xu$^{*,1}$}
\author[ ]{Tian Liang$^{*,1}$}
\author[ ]{Zhiwei He$^{*,1,2}$}
\author[ ]{Jianhui Pang$^{1}$}
\author[ ]{Dian Yu$^{1}$}
\author[ ]{\mbox{Linfeng Song}$^{1}$}
\author[ ]{Qiuzhi Liu$^{1}$}
\author[ ]{Mengfei Zhou$^2$}
\author[ ]{Zhuosheng Zhang$^2$}
\author[ ]{Rui Wang\thanks{Correspondence to: Zhaopeng Tu \textless zptu@tencent.com\textgreater ~and Rui Wang \textless wangrui12@sjtu.edu.cn\textgreater.}~~$^2$}
\author[ ]{\mbox{Zhaopeng Tu}$^{\dag 1}$}
\author[ ]{Haitao Mi$^{1}$}
\author[ ]{Dong Yu$^{1}$}

\affil[1]{Tencent AI Lab}
\affil[2]{Shanghai Jiao Tong University}

\begin{document}

\maketitle

\begin{figure}[h!]
\centering
\vspace{-10mm}
\subfigure[Generated tokens on question ``what is the answer of 2 plus 3?'']{\includegraphics[width=0.46\linewidth]{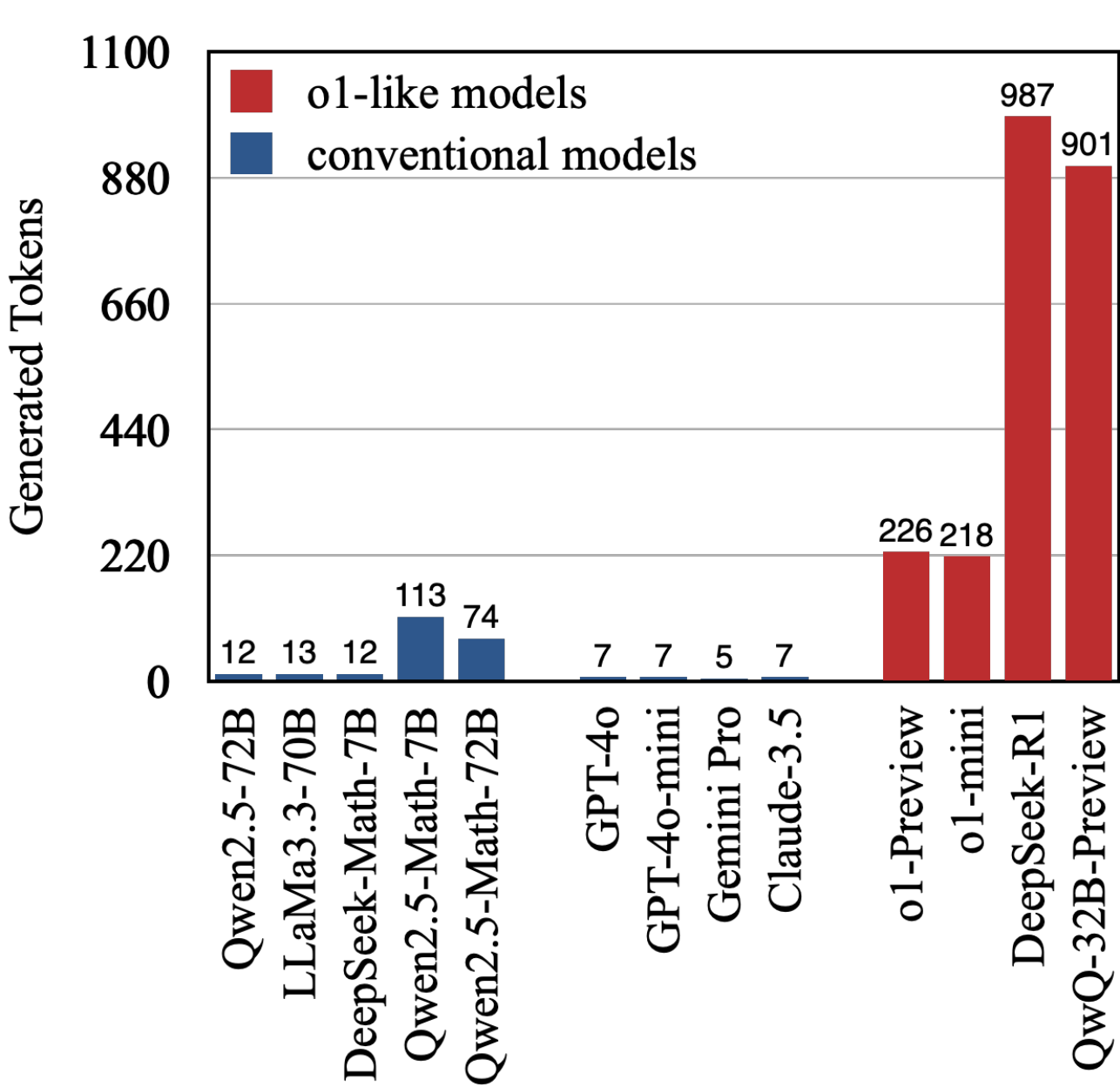}}
\hfill
\subfigure[Token-accuracy plot on MATH500]{\includegraphics[width=0.48\linewidth]{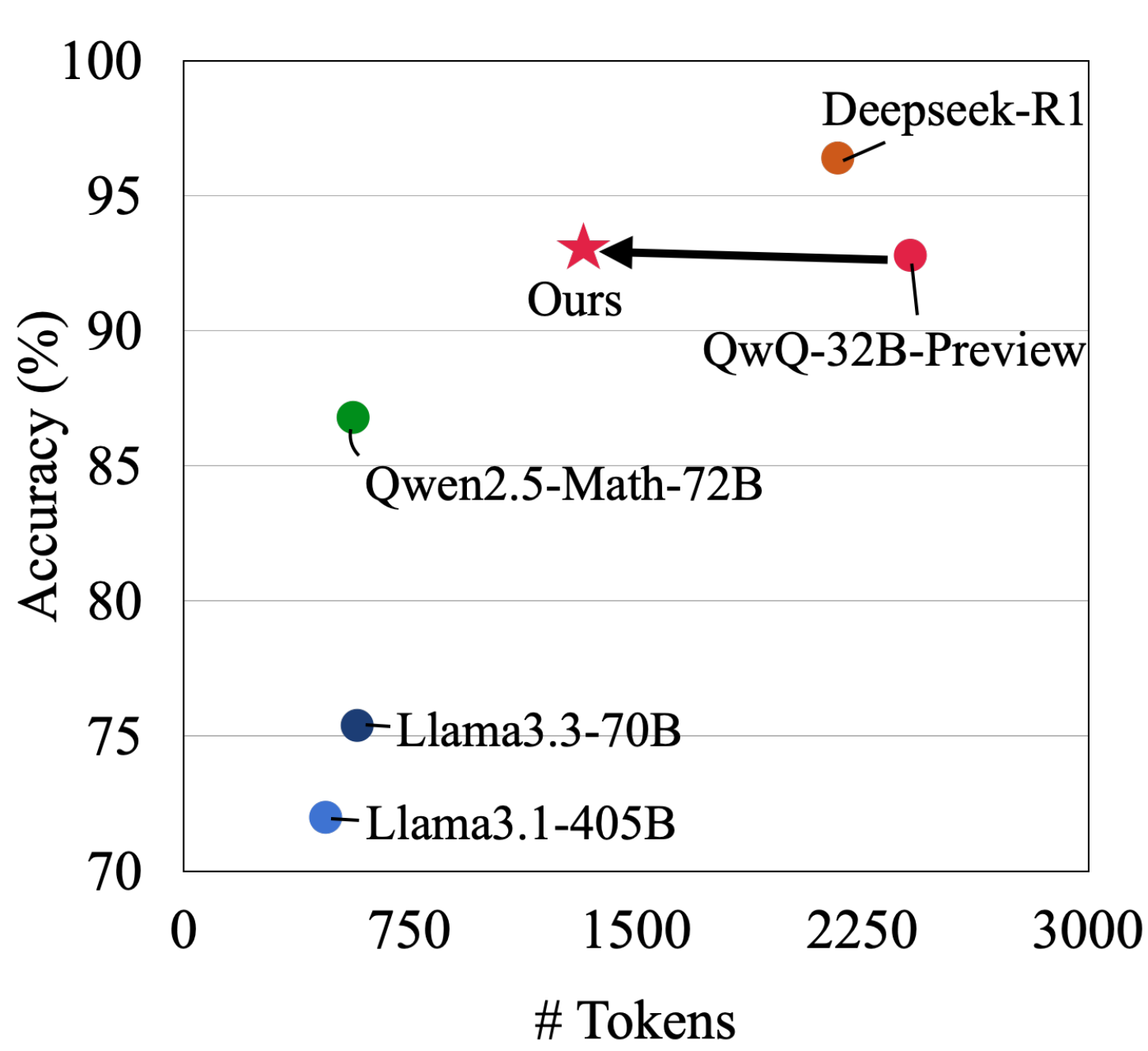}}
\caption{Illustration of {\bf overthinking issue} in Figure (a): o1-like models (right panel) spend much more tokens than conventional LLMs (left and middle panels). Our method reduces the overthinking issue when applied to QwQ-32B-Preview (Figure (b)).}
\label{fig:overthinking}
\end{figure} 

\begin{abstract}
The remarkable performance of models like the OpenAI o1 can be attributed to their ability to emulate human-like long-time thinking during inference. These models employ extended chain-of-thought (CoT) processes, exploring multiple strategies to enhance problem-solving capabilities. 
However, a critical question remains: {\em How to intelligently and efficiently scale computational resources during testing}. This paper presents the first comprehensive study on the prevalent issue of {\bf overthinking} in these models, where excessive computational resources are allocated for simple problems with minimal benefit. We introduce novel efficiency metrics from both outcome and process perspectives to evaluate the rational use of computational resources by o1-like models. Using a self-training paradigm, we propose strategies to mitigate overthinking, streamlining reasoning processes without compromising accuracy. 
Experimental results show that our approach successfully reduces computational overhead while preserving model performance across a range of testsets with varying difficulty levels, such as GSM8K, MATH500, GPQA, and AIME.
\end{abstract}

\section{Introduction}

The OpenAI o1 model~\citep{openai-learning-to-reason} and its replicas~\citep{qwq-32b-preview,guo2025deepseek,team2025kimi} exemplify the state-of-the-art in AI reasoning. 
Their success is largely attributed to mimicking human-like long-time thinking before responding to a question.
Specifically, o1-like models cultivate a long chain-of-thoughts (CoT), explore multiple strategies, break down complex steps, and perform double-checking, which ultimately enhance their ability to tackle intricate reasoning tasks. This approach, known as ``scaling test-time compute'', involves allocating more computational resources during the model's inference phase to generally yield more accurate responses.

\begin{figure*}[t!]
    \centering
    \includegraphics[width=\linewidth]{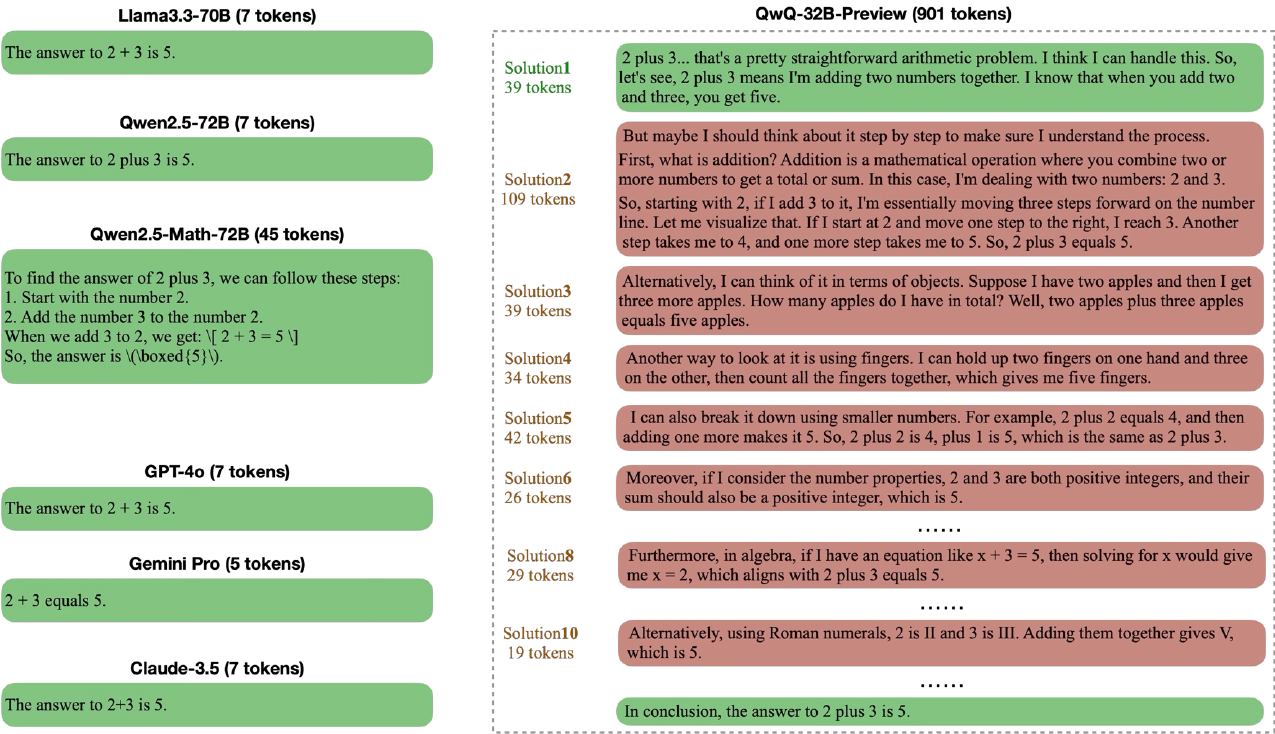}
    \caption{An example of overthinking issue for QwQ-32B-Preview model's output response that consists of 13 solutions. We also list the outputs of other conventional LLMs for reference.}
    \label{fig:overthinking_case}
\end{figure*}

While effective, a critical yet underexplored question remains: {\bf Are we scaling test-time compute efficiently and intelligently?} This study provides an initial exploration of this problem. We first observe that o1-like models exhibit significant {\bf overthinking} issues. Specifically, they tend to expend excessive compute (in terms of tokens or thinking rounds) on questions that are exceptionally simple or for which the answer is already evident. For example, Figure~\ref{fig:overthinking}(a) compares the token usage of o1-like models with conventional models when answering the question, ``what is the answer of 2 plus 3?'' 
On average, o1-like models consumed {\bf 1,953\%} more tokens than conventional models to reach the same answer. Figure~\ref{fig:overthinking_case} illustrates a concrete example where o1-style thinking results in generating 13 solutions for this trivially simple question. Across extensive analyses of mathematical benchmarks, we found these overthinking patterns: {(1)} contribute minimally to improving accuracy, {(2)} lack diversity in reasoning strategies, and {(3)} occur more frequently with simple problems.

The overthinking observed in o1-like models reveals inefficiency in inference and highlights fundamental limitations in their reasoning and decision-making processes. We assert that reasoning involves not only accuracy but also the application of the appropriate level of complexity based on the problem's requirements. This insight motivates our exploration of studying and mitigating overthinking. To address this, we propose two metrics from both outcome and process perspectives to evaluate o1-like models' efficiency. These metrics help provide a comprehensive assessment of the {\bf efficiency} of o1-like models, augmenting the commonly-used {\bf effectiveness} metrics.

To mitigate overthinking without introducing external information, we adopt a self-training paradigm. With our proposed efficiency metrics, we streamline the generated responses by removing redundant solutions while maintaining basic reflexivity. Experimental results across testsets of varying difficulty levels (e.g., GSM8K, MATH500, GPQA, and AIME) demonstrate our approach's effectiveness and robustness in mitigating overthinking issues. For instance, as shown in Figure~\ref{fig:overthinking}(b), our approach can reduce token output by 48.6\% while maintaining accuracy on the widely-used MATH500 testset as applied to QwQ-32B-Preview.

In summary, our contributions are three-fold:
\begin{enumerate}[leftmargin=10pt]
    \item We present the first study offering both a definitive explanation and comprehensive analysis of the overthinking issue, showing that o1-like LLMs often expend unnecessary computational resources on redundant solutions that contribute minimally to final outcomes.
    \item We introduce metrics considering both outcome and process aspects to assess the efficiency of o1-like models.
    \item We explore several strategies to mitigate overthinking, significantly reducing token generation while maintaining model performance across testsets of varying difficulty.
\end{enumerate}

\section{Observing Overthinking Issues}
\label{sec:observing_overthinking}

In this section, we present a comprehensive analysis of outputs generated by o1-like models. First, we provide a basic illustration of the solution distribution in responses from these models (\S~\ref{sec:solution_distribution}). We then identify two inefficiencies in long CoT responses: their limited contribution to accuracy (\S~\ref{sec:solution_contribution}) and diversity (\S~\ref{sec:solution_diversity}). To evaluate these inefficiencies empirically, we propose two efficiency metrics based on our observations. Finally, we present empirical results in \S~\ref{sec:efficiency_results} and conclude that {\em o1-like models often overthink, particularly with easier math problems}.

\subsection{Solution Distribution of o1-Like Models}
\label{sec:solution_distribution}

\paragraph{Experimental Setup} 
We conduct experiments on three testsets: 
\begin{itemize}[leftmargin=10pt]
    \item {\bf ASDIV}~\citep{miao-etal-2020-diverse}: an English math word problem corpus with 2,305 instances, each annotated with its problem type and grade level (1 to 6, indicating difficulty). The test set covers three main problem types (i.e., {\em basic arithmetic operations}, {\em aggregative operations}, and {\em additional domain knowledge required}), typically found in {\bf elementary schools}.
    \item {\bf GSM8K}~\citep{gsm8k}: a dataset of high-quality, linguistically diverse {\bf grade school math word problems} created by human problem writers. The test set includes 1,319 problems, with solutions often involving a sequence of elementary calculations using basic arithmetic. A middle school student should be able to solve every problem.
    \item {\bf MATH500}~\citep{hendrycks2021measuring}: a challenging dataset consisting of problems from {\bf high school math competitions} across seven subjects (e.g., Prealgebra, Algebra, Number Theory) and difficulty levels based on AoPS (ranging from 1 to 5). Problems in these competitions range from level 1, the easiest, often found in AMC 8 exams, to level 5, like those in AIME.
\end{itemize}
The overall difficulty levels of the test sets are ASDIV $<$ GSM8K $<$ MATH500.

We mainly investigate two widely recognized o1-like models featuring a visible thinking process: Qwen-QwQ-32B-Preview~\citep{qwq-32b-preview} and DeepSeek-R1~\citep{DeepSeekAI2025DeepSeekR1IR}.

\begin{figure}[h!]
\centering
    \includegraphics[width=0.8\textwidth]{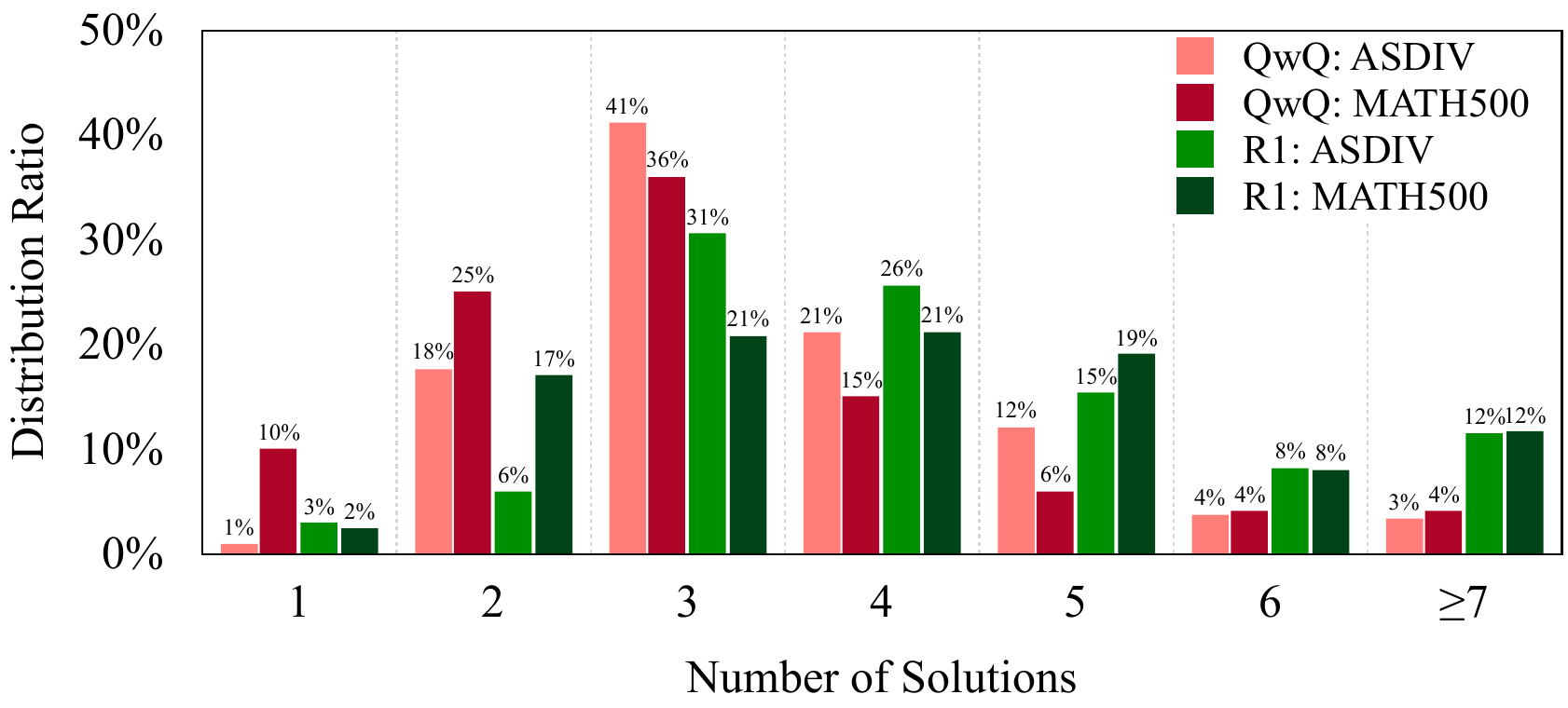}
    \caption{Distribution of solution counts in generated responses for different test sets and models (QwQ-32B-Preview (``QwQ'') and DeepSeek-R1 (``R1'')).}
    \label{fig:solution_distribution}
\end{figure}

\paragraph{Solution Distribution}
In this paper, we define \textit{solution} as part of the full model generation that contains an answer explicitly. For example, in Figure \ref{fig:overthinking_case}, each solution in the QwQ generation contains the answer $5$. We use the Llama-3.3-70B model to separate solutions from generated responses. Figure~\ref{fig:solution_distribution} shows the distribution of solutions in generated responses. Generally, o1-like models produce 2 to 4 solution rounds for most instances, covering 76\% to 80\% of cases for QwQ-32B-Preview across the test sets and 59\% to 63\% for DeepSeek-R1. Regarding different test sets, o1-like models tend to generate more solutions for easier test sets. 
For instance, the average number of solutions of QwQ-32B-Preview on the easiest ASDIV test set is 3.5, whereas on the most difficult MATH500 test set, it is 3.2. The numbers for DeepSeek-R1 are respectively 4.5 and 4.3.

\begin{figure}[h!]
\centering
    \subfigure[QwQ-32B-Preview]{
    \includegraphics[width=0.35\textwidth]{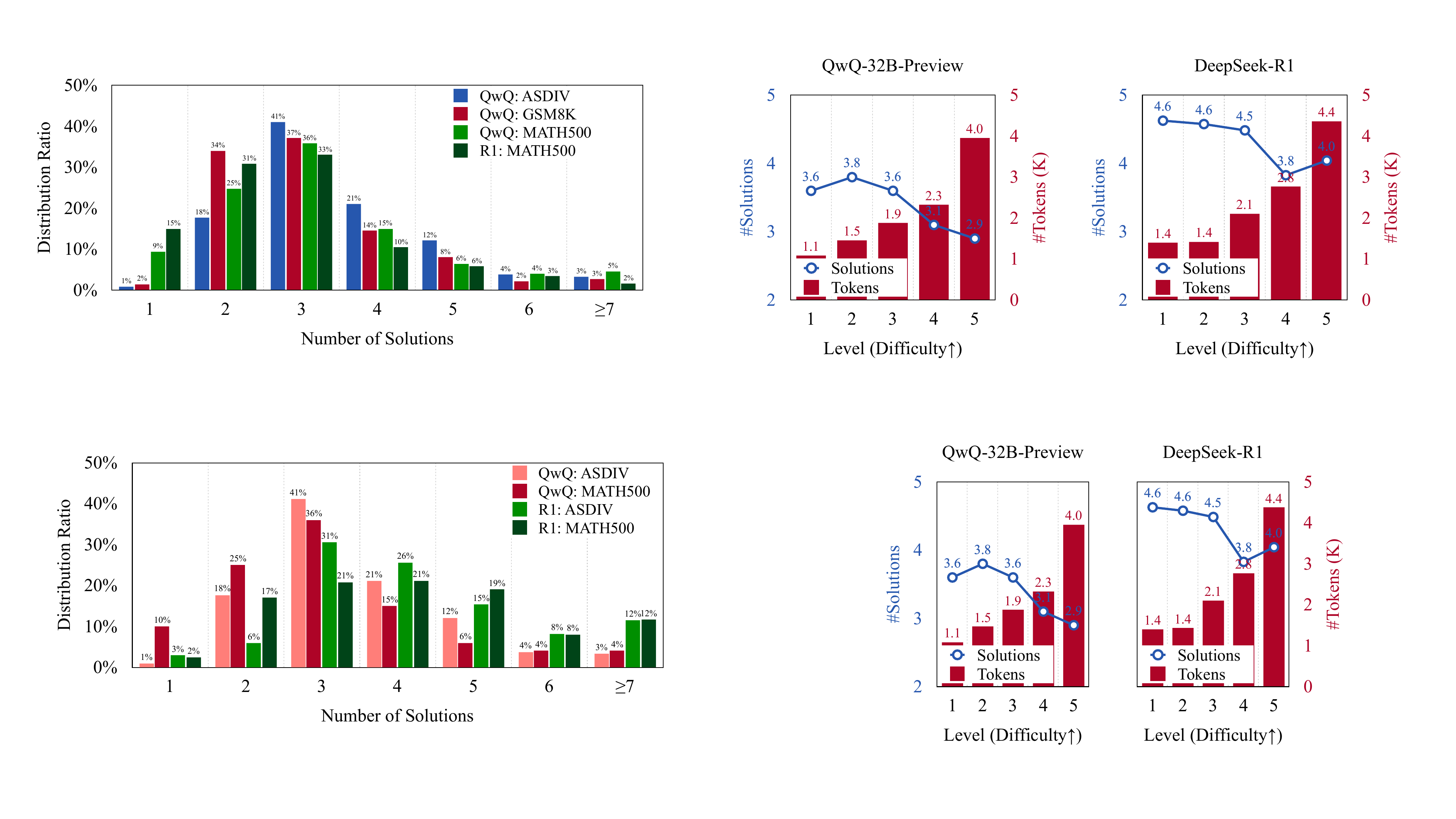}} \hspace{0.1\textwidth}
    \subfigure[DeepSeek-R1]{
    \includegraphics[width=0.35\textwidth]{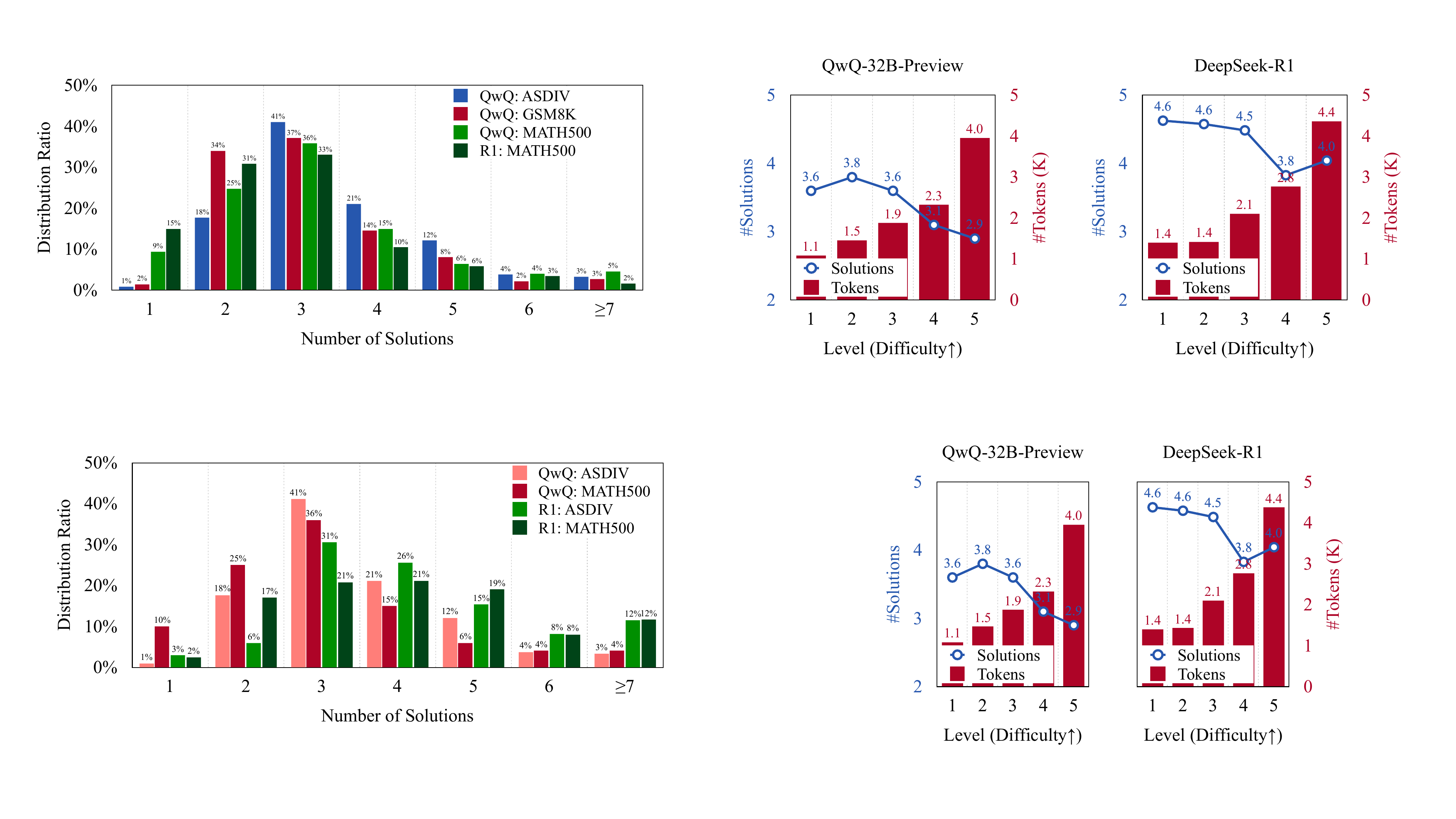}}
    \caption{Average rounds of solutions (``Solutions'') and number of tokens (``Tokens'') in generated responses across different difficulty levels of the MATH500 test set.}
    \label{fig:solution_distribution_math500}
\end{figure}

To empirically validate this finding, we conducted an analysis across various difficulty levels in the MATH500 test set, as illustrated in Figure~\ref{fig:solution_distribution_math500}. Both QwQ-32B-Preview and DeepSeek-R1 generate more solution rounds for problems at easier levels 1-2 (e.g., averaging 3.7 rounds and 4.6 rounds, respectively) compared to levels 4-5 (e.g., averaging 3.0 rounds and 3.9 rounds, respectively), despite the number of tokens consistently increasing with the difficulty level.
These results support our claim that {\bf o1-like models tend to generate more solution rounds for easier math problems}.

\subsection{Efficiency on Accuracy Improvements}
\label{sec:solution_contribution}

\paragraph{Intuition} 
In the example in Figure~\ref{fig:overthinking_case}, we observe that the initial round of solutions already yields the correct answer. Subsequent solutions, which account for the majority of generated tokens, do not enhance accuracy. Based on this observation, we empirically investigate whether later solutions contribute to accuracy improvements. Specifically, for all cases where o1-like models produce the correct answer in the response, we calculate the distribution of occurrences for the first correct answer, termed the ``first correctness distribution''. If more correct answers appear in earlier solutions, then the subsequent solutions contribute minimally to accuracy improvement, indicating reduced efficiency.

\begin{wrapfigure}{r}{0.48\linewidth}
\vspace{-20pt}
\includegraphics[width=\linewidth]{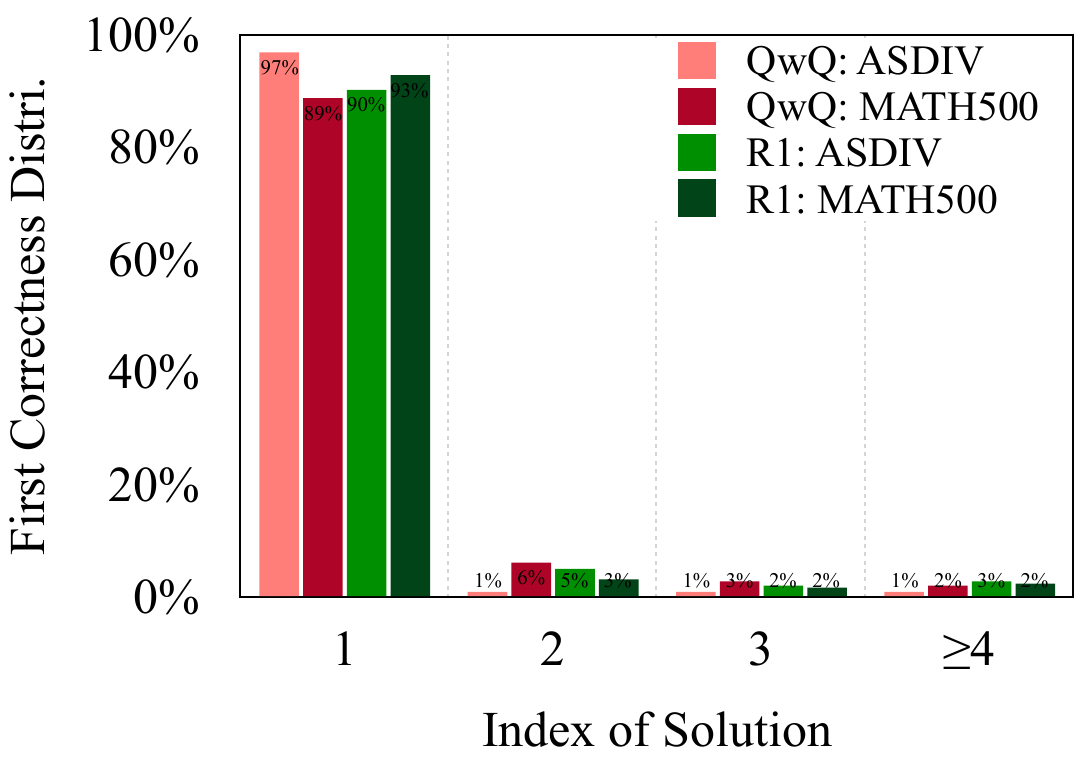}
\caption{Distribution of occurrences for the first correct answer.} 
\label{fig:solution_contribution}
\vspace{-20pt}
\end{wrapfigure}
\paragraph{Observation}
Figure~\ref{fig:solution_contribution} illustrates the first correctness distribution across the test sets and models. In more than 92\% of cases, the initial round of solutions produces the correct answer. Notably, the first round generally comprises less than 60\% of the total tokens generated, suggesting that the extended CoT might not significantly enhance accuracy. For instance, the average length of the first round of solutions for QwQ-32B-Preview on the ASDIV test set is 287 tokens, constituting only 38.7\% of the entire response. These results suggest that {\bf later solutions marginally contribute to improvements in accuracy}.

\paragraph{Outcome Efficiency Metric}
Based on the above observation, we propose an outcome efficiency metric to empirically evaluate how effectively later solutions contribute to accuracy improvements. The outcome efficiency metric, denoted \(\xi_O\), is defined by the following formula:
\begin{equation}
    \xi_O = \frac{1}{N} \sum_{i=1}^N \sigma_i \frac{\hat{T}_i}{T_i}
    \label{eq:outcome_efficiency}
\end{equation}  
where \(N\) is the number of instances in a given test set, $T_i$ is the total number of tokens produced for the $i$-th instance, and $\hat{T}_i$ denotes the efficient tokens that contribute to reaching the correct answer:
\[
\hat{T}_i =  
\begin{cases}  
\text{\#tokens to first arrive at correct answer}, & \sigma_i = 1\\  
T_i, & \sigma_i = 0 
\end{cases}  
\]

$\sigma_i$ denotes whether the evaluated model can produce a correct answer in the response:
\[
\sigma_i =  
\begin{cases}  
1, & \text{if at least one solution in response is correct} \\  
0, & \text{otherwise}  
\end{cases}  
\]
Intuitively, if a model correctly answers at an early stage, the tokens generated thereafter do not contribute to improving accuracy and are considered inefficient. 
% Conversely, if the model fails to produce a correct answer, then all tokens are considered efficient in the attempt to reach correctness, despite failure.
Consider Figure~\ref{fig:overthinking_case} as an example: The first solution correctly addresses the problem with \(\hat{T} = 39\). Consequently, \(\xi_O = \frac{39}{901} = 4.3\%\), which can be considered extremely inefficient.

\iffalse
We can also propose a stricter version of \(\xi_O\), denoted as \(\xi_O^+\), calculated as:
\begin{equation}
    \xi_O^+ = \frac{1}{N} \sum_{i=1}^N \sigma_i \frac{\hat{T}_i}{T_i}
    \label{eq:positive_outcome_efficiency}
\end{equation}
Here only the tokens leading to a correct answer (\(\sigma_i = 1\)) are considered efficient. For instance, since the first solution in Figure~\ref{fig:overthinking_case} is correct, \(\xi_O^+ = \xi_O = 4.3\%\).
 \fi

\subsection{Efficiency on Diverse Thinking}
\label{sec:solution_diversity}

\paragraph{Intuition} 
Some researchers might argue that while solving an easy math problem may appear straightforward, approaching it from different perspectives can deepen understanding and build flexibility in mathematical thinking, which is also valuable. Consider the example output of QwQ-32B-Preview in Figure~\ref{fig:overthinking_case}: Solution 1 states the basic fact that 2 plus 3 equals 5; Solution 2 breaks the addition into smaller steps; Solution 3 uses a counting objects analogy. These three solutions provide different reasoning strategies. However, Solution 4 repeats Solution 3, and Solution 5 repeats Solution 2 using similar reasoning strategies. In this section, we empirically examine the diversity among solutions within a response.

\paragraph{Observation} 
To empirically evaluate whether later solutions provide new reasoning strategies, we introduce the ``distinctness ratio'' as the measure for the ratio of distinct solutions for each data index. 
Consider \(R_i = \{s_i^{1}, \dots, s_i^{m}, \dots, s_i^{M_i}\}\) as the set of \(M_i\) solutions in the \(i\)-th instance response.
Let \(S^m = \{s_1^m, \dots, s_k^m, \dots, s_K^m\}\) be the set of $m$-th solutions in the responses of all instances in the test subset.\footnote{If a response does not contain the $m$-th solution (i.e. \(M_i \textless m\)), that response is excluded from the set, hence $K$ does not necessarily equal the number of test set instances $N$.} The distinctness ratio is defined as:

\[
\mathrm{Dis}^m = \frac{\sum_{k=1}^K \tau_k^m}{K}
\]
where
\[
\tau_k^m =
\begin{cases}
1,  & \text{if } \Phi(s_k^m) \nsubseteq \{\Phi(s_k^1), \dots, \Phi(s_k^{m-1})\} \\
0, & \text{otherwise}
\end{cases}
\]
In this context, \(\Phi(s_k^m)\) is the reasoning strategy of \(s_k^m\). 
We use GPT-4o to cluster the solutions for each instance into groups via a prompt like \citep{ye2024assessing}.\footnote{Refer to Appendix~\ref{app:cluster_prompt} for clustering prompt details.} The clustering results for the QwQ-32B-Preview response in Figure~\ref{fig:overthinking_case} are:

\noindent\fbox{\begin{minipage}{0.98\linewidth}
cluster1   [Solution 1, Solution 6, Solution 11]   stating or affirming the basic arithmetic fact that 2 plus 3 equals 5.\\
cluster2    [Solution 2, Solution5] breaking down the addition into smaller, simpler steps to reach the result.\\
cluster3    [Solution 3, Solution 4]    using a practical analogy of counting objects to explain the addition.\\
cluster4    [Solution 7]    using subtraction as a reverse check to verify the addition result.\\
cluster5    [Solution 8]    using algebraic manipulation and solving simple equations to confirm the result.\\
cluster6    [Solution 9, Solution 10]   converting numbers into different systems (binary and Roman numerals) to verify the result.\\
cluster7    [Solution 12, Solution 13]  considering specific contexts or frameworks like modular arithmetic or programming which could change traditional addition results.
\end{minipage}}

\vspace{5pt}

\begin{wrapfigure}{r}{0.48\linewidth}
\vspace{-15pt}
\centering
    \includegraphics[width=\linewidth]{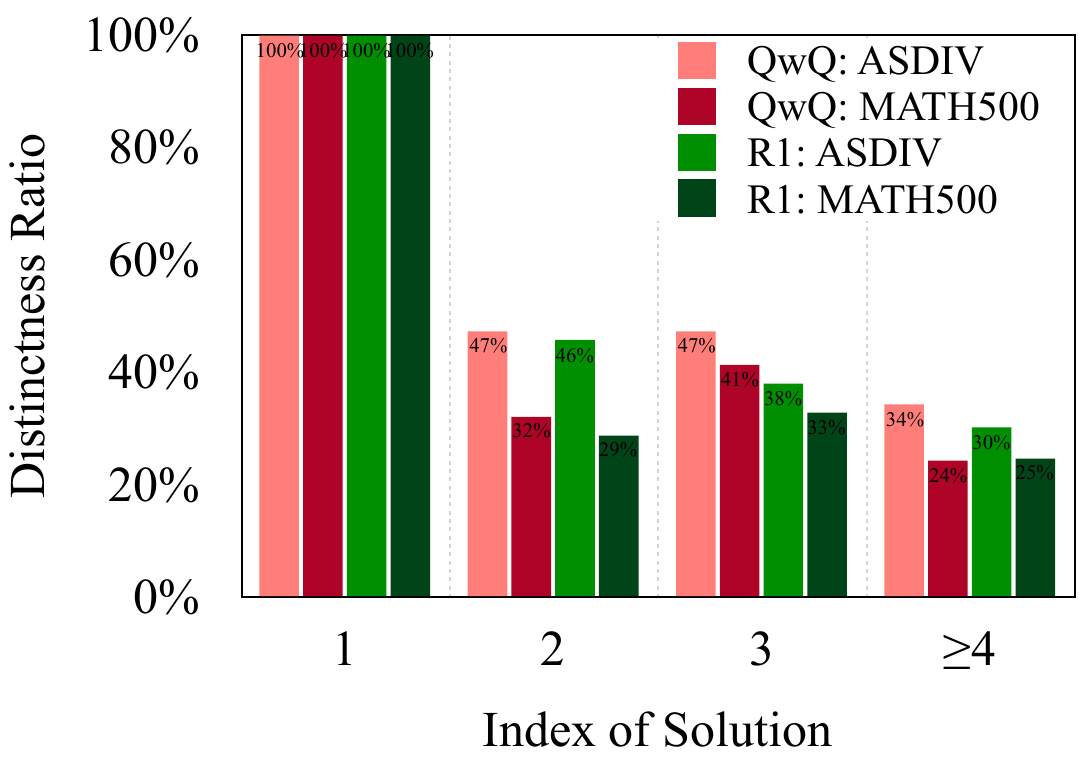}
    \caption{Ratio of whether a solution provides a new reasoning strategy for each index.} % {\bf Later solutions tend to repeat earlier ones.}}
    \label{fig:solution_diversity}
    \vspace{-20pt}
\end{wrapfigure}
Figure~\ref{fig:solution_diversity} displays the distinctness ratio for each solution index. Intuitively, the ratio for Solution\#1 is always 100\%, as it has no preceding solutions, thus \(\tau \equiv 1\) for all instances. Generally, the ratio decreases with higher indices, indicating that {\bf later solutions often repeat earlier ones}. For example, the average distinctness ratio for Solution\#$\geq$4 across test sets decreases by 11.5\% compared to Solution\#3. The ratio of Solution\#2 significantly decreases, underperforming Solution\#3. By reviewing outputs, we find that Solution\#2 often double-checks answers from Solution\#1 using the same reasoning strategy. Subsequently, Solution\#3 attempts to solve the problem using a new reasoning strategy.

% \paragraph{Later Solutions Tend to Repeat Earlier Ones.}

\paragraph{Process Efficiency Metric} 
Based on the above observation, we propose a process efficiency metric to empirically evaluate the contribution of later solutions to solution diversity. The process efficiency metric, denoted \(\xi_P\), is calculated using the formula:
\begin{equation}
    \xi_P = \frac{1}{N} \sum_{i=1}^N \frac{D_i}{T_i} \label{eq: process_efficiency}
\end{equation}
where \(D_i\) represents the number of efficient tokens that contribute to the solutions' diversity. 
Here, we intentionally exclude the factor \(\sigma_i\) to concentrate on diversity, {\bf independent of correctness}.
Let \(T_i^m\) denote the number of tokens in solution \(s_i^m\). We define:
\[
D_i = \sum_{m=1}^M \tau_i^m T_i^m 
\]
\iffalse
where
\[
\tau_{i,m} =
\begin{cases}
1,  & \text{if } \Phi(s_{1,m}) \nsubseteq \{\Phi(s_{i,1}), \dots, \Phi(s_{i,m-1})\} \\
0, & \text{otherwise}
\end{cases}
\]
In this context, \(\Phi(s_{1,m})\) is the reasoning strategy of \(s_{1,m}\).
\fi
Intuitively, the tokens in a distinct solution are regarded as process efficient tokens.
\iffalse
Similarly, we define a stricter version by accounting for the correctness of each solution:
\begin{equation}
    \xi_P^+ = \frac{1}{N} \sum_{i=1}^N \frac{D_i^+}{T_i} = \frac{1}{N} \sum_{i=1}^N \frac{\sum_{m=1}^M \sigma_{i,m} \tau_{i,m} T_{i,m} }{T_i} \label{eq: positive_process_efficiency}
\end{equation}
where \(\sigma_{i,m} = 1\) if the solution \(s_{i,m}\) is correct and \(\sigma_{i,m} = 0\) otherwise.
\fi
In the example shown in Figure~\ref{fig:overthinking_case}, the 13 solutions are categorized into 7 distinct reasoning strategies. Consequently, tokens in Solutions 1, 2, 3, 7, 8, 9, and 12 are efficient, resulting in \(\xi_P = \frac{(39 + 109 + 39 + 29 + 29 + 19 + 59)}{901} = 35.8\%\).

\subsection{Empirical Efficiency Results}
\label{sec:efficiency_results}

\begin{table*}
\centering
\caption{Model efficiency results of strong LLMs.}
\label{tab:efficiency_main_results}
\begin{tabular}{lc rr rr} 
\toprule
\multirow{2}{*}{\textbf{Models}} &   \multirow{2}{*}{\bf  Accuracy}  &  \multicolumn{2}{c}{\bf Response}     &   \multicolumn{2}{c}{\bf Efficiency}\\
\cmidrule(lr){3-4}\cmidrule(lr){5-6}
    &   &  \bf \#Solution   &  \bf \#Token   &  \bf Outcome & \bf Process\\
\midrule
\multicolumn{6}{c}{\bf \em ASDIV}\\
Llama-3.3-70B-Instruct & 95.6 & 1.0 & 166.4 & 95.6\% & 100.0\% \\
Qwen2.5-Math-72B-Instruct & 96.3 & 1.0 & 213.0 & 96.3\% & 100.0\% \\
\hdashline
QwQ-32B-Preview & 96.9 & 3.5 & 741.8 & 41.9\% & 66.5\% \\
DeepSeek-R1 & 97.1 & 4.5 & 845.0 & 45.9	\% & 64.3\% \\
\midrule
\multicolumn{6}{c}{\bf \em GSM8K}\\
Llama-3.3-70B-Instruct & 92.6 & 1.0 & 220.3 & 92.6\% & 100.0\% \\
Qwen2.5-Math-72B-Instruct & 95.8 & 1.0 & 317.4 & 95.8\% & 100.0\% \\
\hdashline
QwQ-32B-Preview & 94.8 & 3.1 & 772.8 & 50.7\% & 67.6\% \\
DeepSeek-R1 & 96.4 & 4.3 & 1056.3 & 48.9\% & 62.0\% \\
\midrule
\multicolumn{6}{c}{\bf \em MATH500}\\
Llama-3.3-70B-Instruct & 75.4 & 1.0 & 553.4 & 75.4\% & 100.0\% \\
Qwen2.5-Math-72B-Instruct & 86.8 & 1.0 & 593.1 & 86.8\% & 100.0\% \\
\hdashline
QwQ-32B-Preview & 93.0 & 3.2 & 2407.9 & 52.3\% & 71.2\% \\
DeepSeek-R1 & 96.4 & 4.3 & 2704.3 & 51.0\% & 66.2\% \\
\bottomrule
\end{tabular}
\end{table*}

Table~\ref{tab:efficiency_main_results} presents the results on model efficiency. For comparison, we include two representative conventional LLMs: Llama-3.3-70B-Instruct and Qwen2.5-Math-72B-Instruct. These conventional LLMs produce only a single solution, meaning that \(\frac{D_i}{T_i} = \frac{\hat{T}_i}{T_i} = 1\). Therefore, in these cases, the outcome efficiency metric \(\xi_O = \frac{1}{N} \sum_{i=1}^N \sigma_i\) equals accuracy, and the process efficiency metric \(\xi_P = 1.0\). In comparison, o1-like models generate significantly longer responses, which are less efficient in improving accuracy and solution diversity. We refer to the inefficient use of generated tokens as the ``{\bf overthinking issue}''.
The experimental results demonstrate that while o1-like models have the capacity to generate multiple solutions, their efficiency is hindered by the overthinking issue.

\begin{figure}[t]
\centering
    \subfigure[QwQ-32B-Preview]{
    \includegraphics[width=0.4\textwidth]{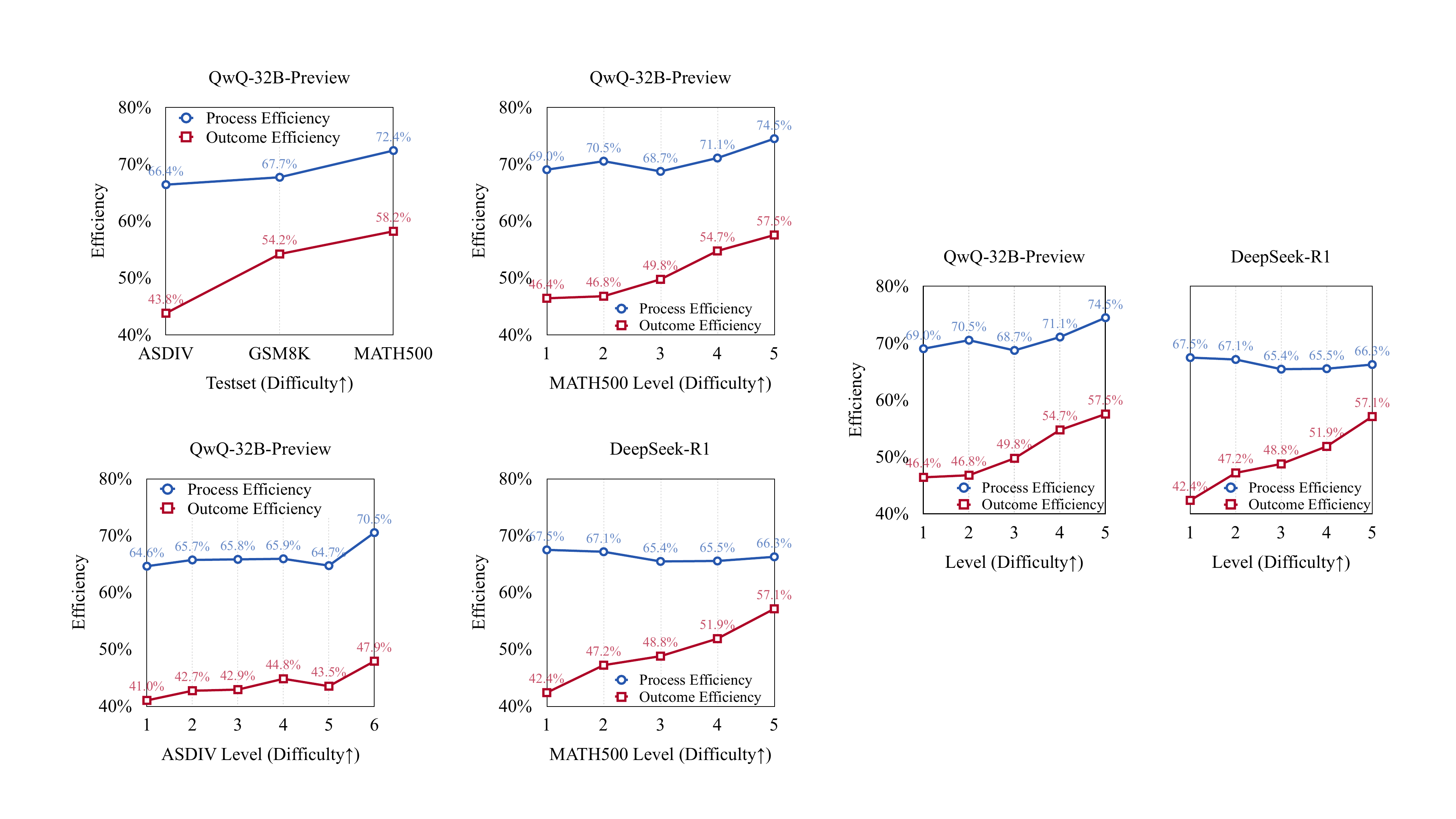}} \hspace{0.1\textwidth}
    \subfigure[DeepSeek-R1]{
    \includegraphics[width=0.4\textwidth]{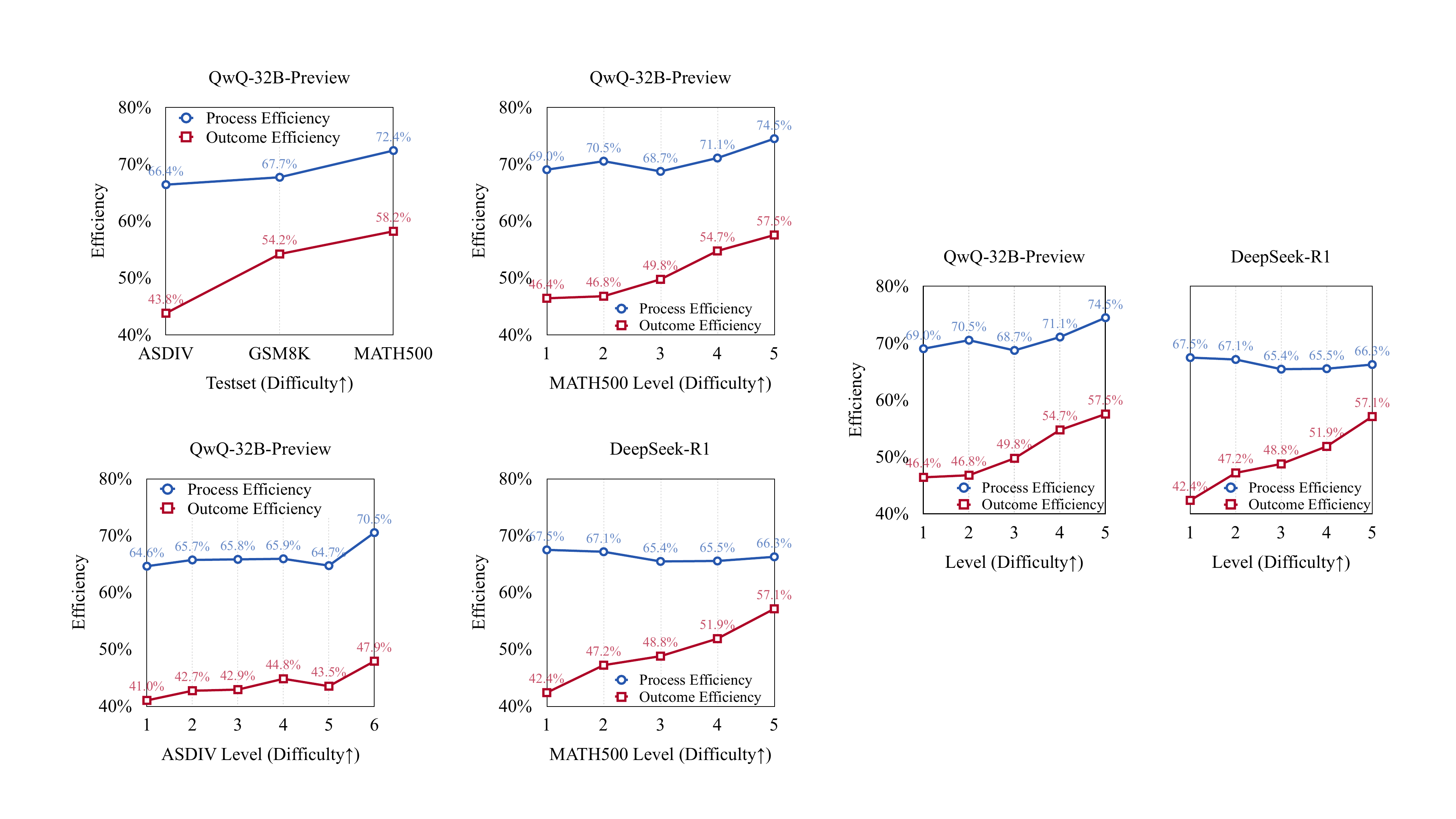}}
    \caption{Efficiency results of (a) QwQ-32B-Preview and (b) DeepSeek-R1 across different difficulty levels of the MATH500 testset.}
    \label{fig:solution_efficiency_math500}
\end{figure}

Figure~\ref{fig:solution_efficiency_math500} presents the detailed efficiency results across various difficulty levels of the MATH500 test set. Notably, both models perform poorly on the simplest Level 1 problems, achieving less than 50\% outcome efficiency, a pattern that corresponds with results observed on the easy ASDIV test set. These findings underscore that {\bf the overthinking issues faced by o1-like models are particularly pronounced with simpler math problems}.

\section{Mitigating Overthinking Issues}

In this section, we explore several strategies aimed at enhancing the efficiency of o1-like models. We adopt the settings for LLM reasoning tasks and primarily utilize the self-training strategy \citep{zelikman2022star,ho2023large}, where the model itself generates the training data. Consistent with previous studies, we employ the PRM12K dataset \citep{lightmanlet} as our training dataset to generate self-training data. The QwQ-32B-Preview model serves as our testing platform because it is available for post-training.

\subsection{Length Preference Optimization}
\label{sec:lpo}

\begin{table}[t]
%\vspace{-10pt}
\centering
\setlength{\tabcolsep}{5pt}
\caption{Statistics on different types of generated responses based on the training data. ``Greedy'' denotes responses generated via greedy search; “Shortest” and “Longest” indicate the shortest and longest responses among 10 samples, respectively.}
\label{tab:training_data_efficiency}
\begin{tabular}{c r r c c} 
\toprule
\multirow{2}{*}{\bf Response}    &   \multirow{2}{*}{\bf \#Solution}  &   \multirow{2}{*}{\bf \#Token} &   \multicolumn{2}{c}{\bf Efficiency}\\
\cmidrule(lr){4-5} 
    &   &   &   \bf Outcome &  \bf Process\\
\midrule
Greedy      &   3.1 & 1434.8    &   55.6\%  &   72.6\%\\  
\hdashline
Shortest    &   2.5 & 1051.3    &   69.8\%  &   80.3\%\\
Longest     &   4.1 & 2258.7    &   46.0\%  &   66.4\%\\
\bottomrule
\end{tabular}
%\vspace{-10pt}
\end{table}

We began by assessing whether the model could produce more efficient responses. We generated 10 samples for each instance in the training dataset with a temperature of 1.0. We discard samples that failed to generate a correct answer. Table~\ref{tab:training_data_efficiency} presents the statistics of different types of generated responses. Our analysis of these sampled responses reveals that the shortest response performs better in terms of both outcome and process efficiency, using fewer rounds and tokens. These findings support our initiative to enhance model efficiency through self-improvement.

We explore several effective methods for self-improvement:
\begin{itemize}[leftmargin=10pt]
    \item {\bf Supervised Fine-Tuning} (SFT;~\citealt{wei2022finetuned}): This method involves fine-tuning a model using positive synthetic data. The model learns to map inputs to preferred outputs by minimizing the cross-entropy loss between predicted and actual outputs. SFT enables the model to mimic the behavior demonstrated in training examples.
    \item {\bf Direct Preference Optimization} (DPO;~\citealt{rafailov2024direct}): This method trains a model directly on human-preferred responses to increase the likelihood of preferred responses over unpreferred ones. % DPO implicitly optimizes the objective by aligning reward functions with optimal policies.
    \item {\bf Reasoning Preference Optimization} (RPO;~\citealt{pang2024iterative,liu2024provably}): This approach modifies the DPO loss by adding a negative log-likelihood term on the preferred response. RPO enhances DPO training stability by % maintaining desired formatting for generated content and 
    preventing a decreased probability of selected responses.
    \item {\bf Simple Preference Optimization} (SimPO;~\citealt{meng2024simpo}): This method addresses the discrepancy between the reward function and the generation metric during inference found in other preference optimization methods. SimPO incorporates techniques like adaptive margin and length regularization into DPO training.
\end{itemize}
Apart from the SFT method, which uses only the shortest sampled response as training data, the other three preference optimization methods require contrastive instance pairs (positive, negative). It is straightforward to use the response generated by greedy search as the negative example, aligning with the real-time inference scenario. However, in our preliminary experiments, we found it less effective than using the longest sampled response as the negative example. One possible reason is that the longest sampled response provides a clearer contrastive signal.

\subsection{Streamlining Responses to Enhance Efficiency}
\label{sec:simres}

\begin{table}[t]
%\vspace{-10pt}
\setcounter{table}{2}
\setlength{\tabcolsep}{4pt}
\centering
\caption{Statistics on different types of positive examples. ``\#S'' denotes the number of solutions.}
\label{tab:response_simplification}
\begin{tabular}{c r r rr} 
\toprule
\multirow{2}{*}{\bf Positive Example}    &   \multirow{2}{*}{\bf \#S}  &   \multirow{2}{*}{\bf \#Token} &   \multicolumn{2}{c}{\bf Efficiency}\\
\cmidrule(lr){4-5} 
    &   &   &   \bf Outcome &  \bf Process\\
\midrule
Shortest Response                       &   2.5 & 1051.3    &   69.8\%  &   80.3\%\\
\hdashline
FCS    &   1.1 & 681.0         &  99.5\%   &  99.1\%\\
FCS + Ref.                &   1.9 & 878.7         &  78.4\%   &  82.4\% \\
GDS        &   1.6 & 856.8         &  86.8\%   &  94.2\%\\
\bottomrule
\end{tabular}
\end{table}

Although shorter sampled responses improve the efficiency of o1-like models, they still suffer from overthinking issues. Based on the observations in Section~\ref{sec:observing_overthinking}, where earlier solutions in the response are more efficient, we further streamline the responses to enhance efficiency. We propose three types of simplification strategies that differ in how they streamline the responses from the beginning:
\begin{itemize}[leftmargin=10pt]
    \item \textbf{First-Correct Solutions (FCS)}: This strategy retains the earliest solutions that first arrive at the correct answer.
    \item \textbf{FCS+Reflection}: Since the majority of responses achieve the correct answer on the first solution (see Figure~\ref{fig:solution_contribution}), maintaining only the First-Correct Solutions might cause o1-like models to revert to conventional LLM behavior. To counter this, we extend the approach to include the second solution that reaches the correct answer in positive examples, recalling the model's long-reflective capability while maintaining efficiency.
    \item \textbf{Greedily Diverse Solutions (GDS)}: Figure~\ref{fig:solution_diversity} demonstrates that the distinctiveness of Solution\#2 significantly decreases because the second solution often double-checks answers from the first using the same reasoning strategy. Consequently, FCS+Reflection may reduce efficiency. To address this issue, we propose a simple heuristic that greedily expands solutions providing new reasoning strategies. Additionally, this strategy includes more solutions when the second solution does not repeat the first, thereby increasing diversity.
\end{itemize}

For each instance, we select the shortest result of each type from 10 samples. 
Consequently, the three types of simplified responses may originate from different original responses. 
Table~\ref{tab:response_simplification} presents the statistics for these simplified responses. Notably, all simplified responses enhance efficiency compared to the shortest response. ``FCS'' is the most efficient, both in terms of outcome and process, using the fewest number of solution rounds and tokens. ``FCS+Reflection'' incorporates reflection, requiring approximately one additional solution round, which reduces both outcome and process efficiencies. ``Greedily Diverse Solutions'' serves as a compromise, balancing the number of solutions and tokens, and achieving moderate to high efficiency.

\subsection{Experimental Results}

\begin{table*}[t]
\setcounter{table}{3}
\centering
\caption{Experimental results of the proposed efficiency enhancing methods.}
\label{tab:main_results}
\begin{tabular}{lc rr rr} 
\toprule
\multirow{2}{*}{\textbf{Methods}} &   \multirow{2}{*}{\bf  Accuracy}  &  \multicolumn{2}{c}{\bf Response}     &   \multicolumn{2}{c}{\bf Efficiency}\\
\cmidrule(lr){3-4}\cmidrule(lr){5-6}
    &   &   \bf \#Solution   &   \bf \#Token   &   \bf Outcome   &  \bf Process\\
\midrule
% Copy Start
\multicolumn{6}{c}{\bf \em ASDIV}\\
QwQ-32B-Preview &96.9 & 3.5 & 741.8 & 41.9\% & 66.5\%  \\
~~~+SimPO$_\text{FCS+Reflection}$ & 96.8 & 2.0 & 381.6 & 77.6\% & 86.0\% \\
\midrule
\multicolumn{6}{c}{\bf \em GSM8K}\\
QwQ-32B-Preview & 94.8 & 3.1 & 772.8 & 50.7\% & 67.6\% \\
~~~+SimPO$_\text{FCS+Reflection}$ & 96.0 & 2.0 & 416.6 & 80.2\% & 87.2\% \\
\midrule
\multicolumn{6}{c}{\bf \em MATH500}\\
QwQ-32B-Preview & 93.0 & 3.2 & 2407.9 & 52.3\% & 71.2\% \\
~~~+SFT$_\text{Shortest Response}$ & 93.2 & 3.0 & 2359.5 & 60.4\% & 75.6\% \\
~~~+DPO$_\text{Shortest Response}$ & 94.0 & 2.7 & 1929.5 & 65.8\% & 79.1\% \\
~~~+RPO$_\text{Shortest Response}$ & 91.6 & 2.7 & 2015.7 & 64.8\% & 79.2\% \\
~~~+SimPO$_\text{Shortest Response}$ & 92.4 & 2.5 & 1871.8 & 67.6\% & 80.9\% \\
\hdashline
~~~+SimPO$_\text{First-Correct Solution}$ & 91.0 & 1.4 & 1016.0 & 88.7\% & 98.1\% \\
~~~+SimPO$_\text{FCS+Reflection}$ (Ours) & 92.8 & 1.9 & 1330.7 & 80.0\% & 89.5\% \\
~~~+SimPO$_\text{Greedily Diverse Solutions}$ & 91.8 & 1.7 & 1286.1 & 84.3\% & 93.6\% \\
\midrule
\multicolumn{6}{c}{\bf \em GPQA}\\
Qwen2.5-Math-72B-Instruct & 46.5 & 1.0 & 811.7 & 46.5\% & 100\% \\
\hdashline
QwQ-32B-Preview & 59.6 & 2.2 & 3228.4 & 51.4\% & 84.3\% \\
~~~+SimPO$_\text{FCS+Reflection}$  & 59.1 & 1.7 & 2085.7 & 55.7\% & 90.4\% \\
\midrule
\multicolumn{6}{c}{\bf \em AIME24}\\
Qwen2.5-Math-72B-Instruct & 23.3 & 1.0 & 1204.5 & 23.3\% & 100.0\% \\
\hdashline
QwQ-32B-Preview & 46.7 & 2.6 & 9480.9 & 38.4\% & 84.4\% \\
~~~+SimPO$_\text{FCS+Reflection}$   & 43.3 & 1.7 & 5154.5 & 39.8\% & 92.0\% \\
% Copy End
\bottomrule
\end{tabular}
\end{table*}

Table~\ref{tab:main_results} presents the results of the proposed methods.
We perform a detailed comparison on MATH500 and validate the most effective approach using the other test sets.
% For each method, we train the model three times using different random seeds (67, 68, and 69) and report the averaged results to draw more robust conclusions.

\paragraph{Performance of Length Preference Optimization Methods}
SFT only slightly reduces the number of solution rounds and tokens compared to the vanilla QwQ-32B-Preview model, underperforming the preference optimization methods. Among these methods, SimPO achieves the best results, reducing the number of generated tokens by 22.3\% on MATH500. Consequently, SimPO is used as the default post-training method in the subsequent experiments.

\paragraph{Performance of Response Simplification Methods}
As anticipated, the First-Correction Solutions strategy achieves the greatest reduction in length. However, this method decreases performance on the difficult MATH500 test set, which may require more rounds of reflection. The ``FCS+Reflection'' approach alleviates this issue and surpasses the FCS method by 1.4\% with an additional round of reflection. The ``Greedily Diverse Solutions'' strategy balances performance with the number of generated tokens. However, it significantly underperforms compared to ``FCS+Reflection'', reinforcing our claim that the difficult MATH500 test set requires the deep inference provided by o1-like models. Hence, we adopt ``FCS+Reflection'' as the default response simplification method.

\paragraph{Results on Challenging Test Sets}
Our approach enhances performance on easier testsets such as ASDIV and GSM8K with fewer tokens, demonstrating the effectiveness and versatility of our method in addressing overthinking issues.
To address the concerns of some researchers that our approach might weaken the ability of o1-like models to tackle complex problems requiring long-term reasoning, we validate our method using more challenging GPQA and AIME test sets. As seen, our approach maintains model performance while using fewer tokens, demonstrating the robustness and generalization capability of our approach.

\section{Related Work}

\subsection{Scaling Test-Time Compute}

Enhancing model performance on complex tasks can be achieved by scaling test-time compute, which involves:

\paragraph{Expanding Search Space} LLMs have strong reasoning abilities, but their auto-regressive decoding often misses optimal solutions. Self-consistency generates multiple responses and use majority voting to select the best answer~\citep{wangself}. Other approaches include best-of-n decoding, minimum Bayes risk decoding~\citep{lightmanlet, li-etal-2023-making, khanovargs, heineman2024improving, wu2024better}, and structured search methods such as Tree-of-Thought, Graph-of-Thought, and Monte Carlo Tree Search~\citep{yao2024tree, besta2024graph, luo2024improve, tian2024toward, wan2024alphazero}.

\paragraph{Human-Like Thinking Patterns} LLMs often use natural language reasoning. Techniques like chain-of-thought encourage step-by-step reasoning instead of direct answers~\citep{wei2022chain, kojima2022large}. This has been expanded with methods like debating, self-correction, self-critique, and plan-and-solve~\citep{liang-etal-2024-encouraging, duimproving, xiong2023examining, kumar2024training, kamoi2024can, ke2023critiquellm, lin-etal-2024-criticbench, yu2024self, wang-etal-2023-plan}. Recent studies also explore latent space reasoning to mimic human cognition~\citep{hao2024traininglargelanguagemodels, goyal2024think}. Advanced models combine these patterns into extensive chains-of-thought, improving accuracy with more reasoning time~\citep{openai-learning-to-reason}.

\subsection{Efficient Thinking}
Scaling the search space and scaling human-like thinking involves two distinct aspects of efficiency: efficient search and efficient thinking.
However, few works specifically focus on efficient thinking in LLMs. \citet{team2025kimi} leveraged the long to short strategy to compress generation context.
\citet{zhao2024automaticcurriculumexpertiteration} encourages the model to terminate reasoning by saying ``I don't know'' when the problem is hard to solve.
\citet{han2024token} introduces token-budget-aware reasoning, where the model is prompted with a specified token budget to guide its reasoning process. There are also several contributions made to predict the distribution of the computation budget and allocate the computation power based on the prompt's difficulty~\citep{damani2024learninghardthinkinputadaptive, wang2024makepennycountdifficultyadaptive,xu-etal-2024-adaption}. Another line of work emphasizes the early stopping strategy to save computation budget while reasoning~\citep{manvi2024adaptiveinferencetimecomputellms, li2024escape}. Moreover, multi-agent framework utilizes large LLMs for difficult tasks while small LLMs for simple tasks~\citep{kirchner2024proververifiergamesimprovelegibility, damani2024learninghardthinkinputadaptive}

\vspace{5pt}
In summary, all the aforementioned works consider conventional models rather than o1-like models with longer chains-of-thought. In contrast, our work first identifies the overthinking problem in o1-like model. Additionally, instead of limiting the reasoning space or leaving the token budget to be specified by the user, we aim to train the model to learn how to think efficiently.

\section{Conclusion}

This study identifies a key challenge in o1-like LLMs —- efficient and intelligent scaling of test-time computational resources. 
We have presented a comprehensive analysis of the overthinking issue in o1-like LLMs.
By highlighting the overthinking phenomenon and proposing efficiency metrics, we enhance our understanding of resource utilization in o1-like models. Our self-training based approach effectively mitigates overthinking, reducing unnecessary computation while maintaining performance across reasoning benchmarks of varying difficulty levels.

This work not only improves model efficiency but also sets the groundwork for future research on optimizing computational resource allocation in AI reasoning tasks. Future directions include exploring adaptive compute strategies that dynamically adjust to problem complexity and refining efficiency metrics for broader model generalization.

\iffalse
\section*{Limitation}
This study has the following limitations:
\begin{itemize}[leftmargin=10pt]
    \item {\bf Model Coverage}: Our analysis of overthinking issues is based solely on two models: QwQ-32B-Preview and DeepSeek-R1. These are the only two o1-like models that exhibit a visible long chain-of-thought (CoT) process. Additionally, our efficiency-enhancing methods are exclusively validated on QwQ-32B-Preview. We plan to validate our findings and approaches on more o1-like models as they become accessible.
    \item {\bf Diversity Measurement}: Our diversity measurement utilizes the powerful GPT-4o for clustering solutions, which is costly and difficult to replicate. In forthcoming revisions, we aim to employ more open-source LLMs for diversity assessment, such as LLama3.3-70B, to facilitate easier replication of the methodology.
    \item {\bf Bias in PRM12K Training Data}: Our current validation relies solely on the widely-used PRM12K dataset, which might compromise the robustness of our approach. We intend to explore larger datasets (e.g., with 1 million instances) to address this concern.
\end{itemize}
This work does not solve the overthinking issue in o1-like models but lays the groundwork for future research.
\fi

\bibliography{main}
\bibliographystyle{colm2024_conference}

\clearpage

\appendix

\section{Appendix}

\subsection{Case Overview for Deepseek-R1-Preview}
\label{app:case_overview_r1}
\begin{figure}[h]
    \centering
    \includegraphics[width=0.8\linewidth]{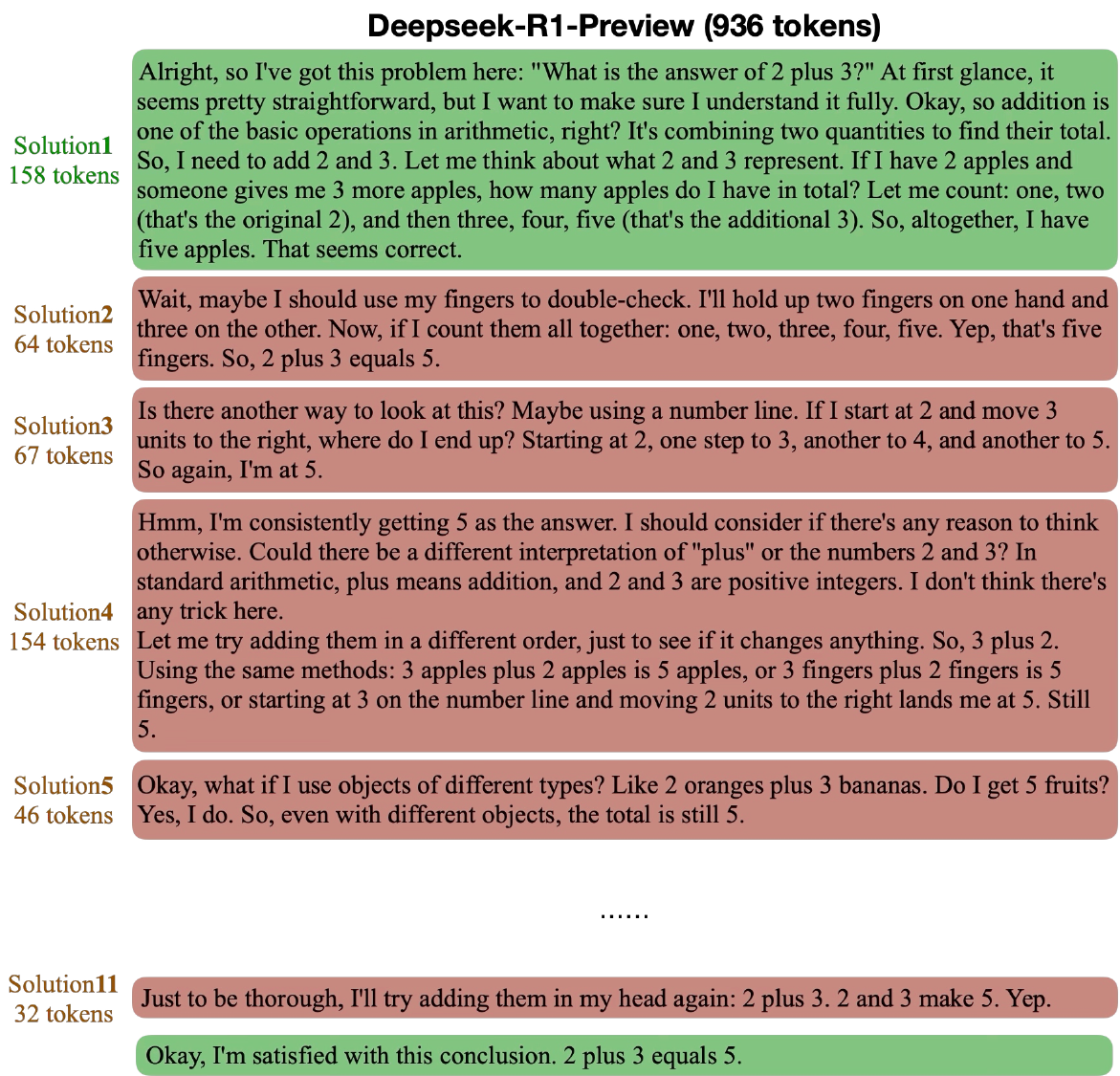}
    \caption{Deepseek-R1-Preview response for the query ``What is the answer of 2 plus 3?''}
    \label{fig:case_overview_r1}
\end{figure}

\subsection{Prompts for Clustering Solutions}
\label{app:cluster_prompt}

Inspired by \citep{ye2024assessing}, we leverage GPT-4o to cluster the solutions for each instance into groups with the following prompt:

\noindent\fbox{\begin{minipage}{0.98\linewidth}
Criteria for clustering the mathematical solutions:\\
1. If the solutions used to arrive at the solutions are fundamentally different from each other, such as algebraic manipulation versus geometric reasoning, they can be considered novel;\\
2. Even if the results are the same, if the intermediate steps or processes involved in reaching those solutions vary significantly, the solutions can be considered different;\\
3. If the solutions relies on different assumptions or conditions, they should be considered different from each other;\\
4. A solution might generalize to a broader class of problems, while another solution might be specific to certain conditions. In such cases, they are considered distinct;\\
5. If one solution is significantly simpler or more complex than the others, it can be regarded as essentially novel, even if they lead to the same result.\\

Given the following mathematical problem:\\
***problem***\\

Solutions:\\
Solution 1: ...\\
Solution 2: ...\\

Please output the clusters strictly following the following format, each row containing a cluster, names, and reasons. Do not include any additional text or explanations outside of this format:

cluster1    [solution names]    reason for cluster\\
cluster2    [solution names]    reason for cluster\\
cluster3    [solution names]    reason for cluster\\
...\\

For example:\\
cluster1    [Solution 1, Solution 3, Solution 5]    similar algebraic approach using the volume formula and canceling terms or directly solving for the height.\\
cluster2    [Solution 2, Solution 4]    verifying the correctness and consistency of the formula and solution and considering unit checks or logical reasoning on how volume relates to base area and height.
\end{minipage}}

\end{document}